\pdfoutput=1

\documentclass[11pt]{article}

\usepackage[final]{acl}
\usepackage{makecell}
\usepackage{times}
\usepackage{latexsym}
\usepackage[most]{tcolorbox}
\usepackage[T1]{fontenc}

\usepackage[utf8]{inputenc}
\usepackage{microtype}

\usepackage{inconsolata}

\usepackage{graphicx}
\usepackage{subfigure}
\usepackage{booktabs} 
\usepackage{hyperref}
\usepackage{multirow} 

\usepackage{amsmath}
\usepackage{enumitem}
\usepackage{amssymb}
\usepackage{mathtools}
\usepackage{amsthm}

\usepackage{amsmath,amsfonts,bm}

\def\eqref#1{equation~\ref{#1}}

\def\1{\bm{1}}

\DeclareMathAlphabet{\mathsfit}{\encodingdefault}{\sfdefault}{m}{sl}
\SetMathAlphabet{\mathsfit}{bold}{\encodingdefault}{\sfdefault}{bx}{n}

\newcommand{\ours}{CaKE~}
\usepackage{wasysym}
\usepackage{marvosym}
\usepackage{todonotes}

\def\is{\leftarrow}

\tcbset{
    mybox/.style={
        colframe=darkgray,          
        colback=darkgray!10,        
        coltitle=white,         
        colbacktitle=darkgray,      
        boxrule=0.5mm,          
        arc=3mm,                
        width=\linewidth,       
        left=1mm,               
        right=1mm,              
        top=1mm,                
        bottom=1mm,             
    }
}

\title{CaKE: Circuit-aware Editing Enables Generalizable Knowledge Learners}

\author{
Yunzhi Yao{$^{\text{\Scorpio\Pisces}}$\thanks{Work done during Yunzhi’s visit to UCLA.}}, Jizhan Fang{$^{\text{\Scorpio}}$}, Jia-Chen Gu{$^{\text{\Pisces}}$}, {\bf Ningyu Zhang}{$^{\text{\Scorpio}}$}\footnotemark[2], \\{\bf Shumin Deng}$^{\text{\Cancer}}$, {\bf Huajun Chen}$^{\text{\Scorpio}}$\thanks{Corresponding Authors}, \textbf{Nanyun Peng}$^{\text{\Pisces}}$\footnotemark[2],\\
$^\text{\Scorpio}$ Zhejiang University 
$^\text{\Cancer}$ National University of Singapore \\
$^\text{\Pisces}$ University of California, Los Angeles\\
  \texttt{\{yyztodd,fangjizhan,zhangningyu,huajunsir\}@zju.edu.cn}\\
  \texttt{gujc@ucla.edu,violetpeng@cs.ucla.edu,shumin@nus.edu.sg}
  }

\begin{document}
\maketitle

\begin{abstract}
Knowledge Editing (KE) enables the modification of outdated or incorrect information in large language models (LLMs). While existing KE methods can update isolated facts, they often fail to generalize these updates to multi-hop reasoning tasks that rely on the modified knowledge. Through an analysis of reasoning circuits---the neural pathways LLMs use for knowledge-based inference, we find that current layer-localized KE approaches (e.g., MEMIT, WISE), which edit only single or a few model layers, inadequately integrate updated knowledge into these reasoning pathways. To address this limitation, we present \textbf{CaKE} (\textbf{C}ircuit-\textbf{a}ware \textbf{K}nowledge \textbf{E}diting), a novel method that enhances the effective integration of updated knowledge in LLMs.  By only leveraging a few curated data samples guided by our circuit-based analysis, CaKE stimulates the model to develop appropriate reasoning circuits for newly incorporated knowledge. Experiments show that CaKE enables more accurate and consistent use of edited knowledge across related reasoning tasks, achieving an average improvement of 20\% in multi-hop reasoning accuracy on the MQuAKE dataset while requiring less memory than existing KE methods.
We release the code and data in \url{https://github.com/zjunlp/CaKE}.
\end{abstract}

\section{Introduction}
\label{sec:introduction}
Large language models (LLMs) have demonstrated remarkable capabilities in diverse tasks~\cite{yang2024qwen2,azaria2024chat,dubey2024llama,openai2021introducing,guo2025deepseek}, achieving performance that rivals or even exceeds human experts.
However, their practical deployment faces some critical limitations: parametric knowledge remains static after pretraining, making it challenging to keep up with evolving real-world information; their propensity for hallucinations also undermines reliability \cite{chen2024large}. 
Knowledge editing (KE) has emerged as a promising solution to update the knowledge in models precisely \cite{mitchell2021fast,wang2024wise,jiang2025anyedit}.
Although existing KE methods achieve good results on simple factual updates~\cite{yao2023editing,zhang2024comprehensive}, they often exhibit fundamental limitations: edits propagate inconsistently through related knowledge structures and downstream reasoning tasks~\cite{cohen2024evaluating,qin2024does,yao2023editing};
excessive focus on surface-level pattern matching~\cite{hoelscher2023detecting}, and locality issues for other unrelated knowledge and general ability~\cite{gu2024model,gupta-etal-2024-model}. \looseness=-1

\begin{figure}
    \centering
    \includegraphics[width=\linewidth]{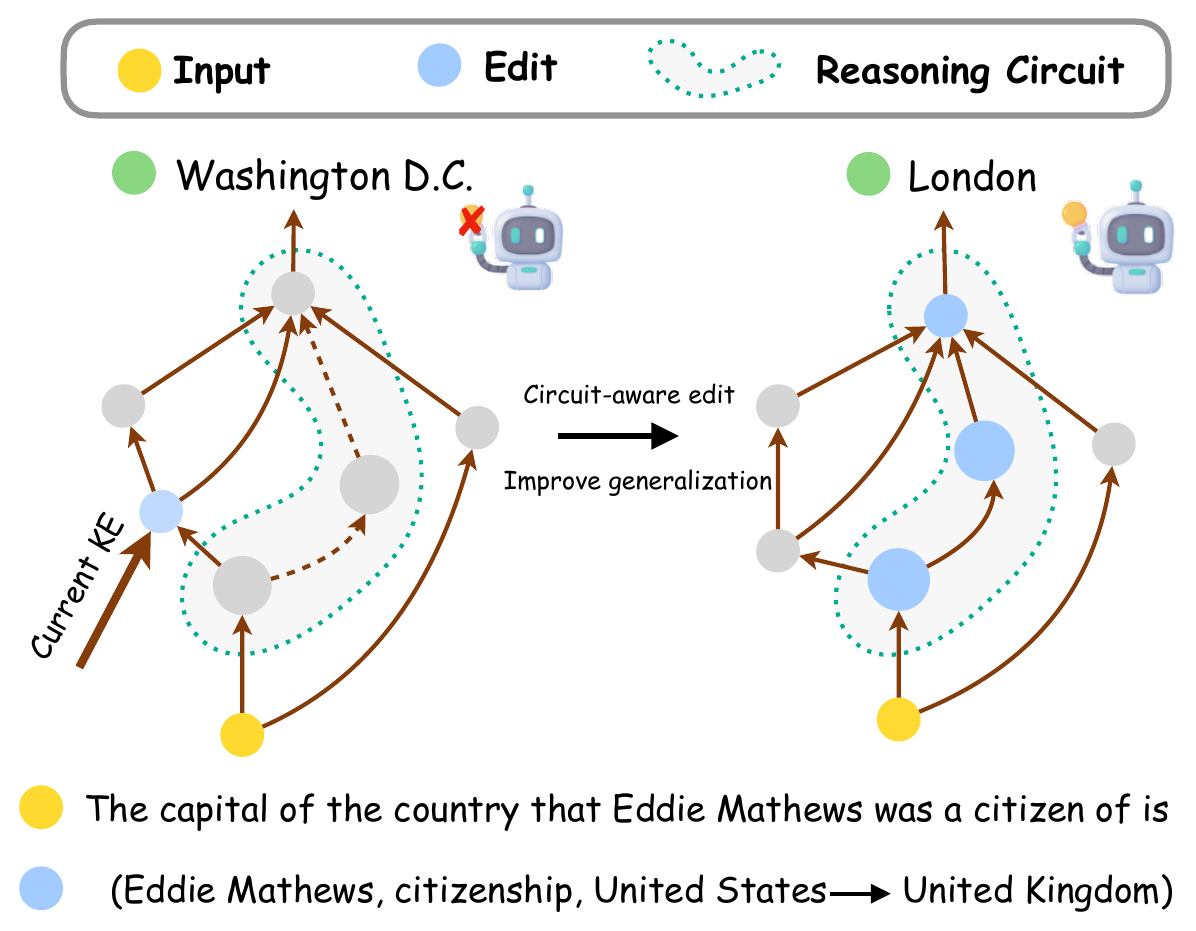}
    \vspace{-10pt}
    \caption{The current edit cannot propagate the new knowledge to the reasoning circuit for multi-hop reasoning. We propose a circuit-aware edit to improve the model's multi-hop reasoning performance involving the updated knowledge.}
    \label{fig:overview}
    \vspace{-10pt}
\end{figure}
\begin{figure*}
\centering
\includegraphics[width=\linewidth]{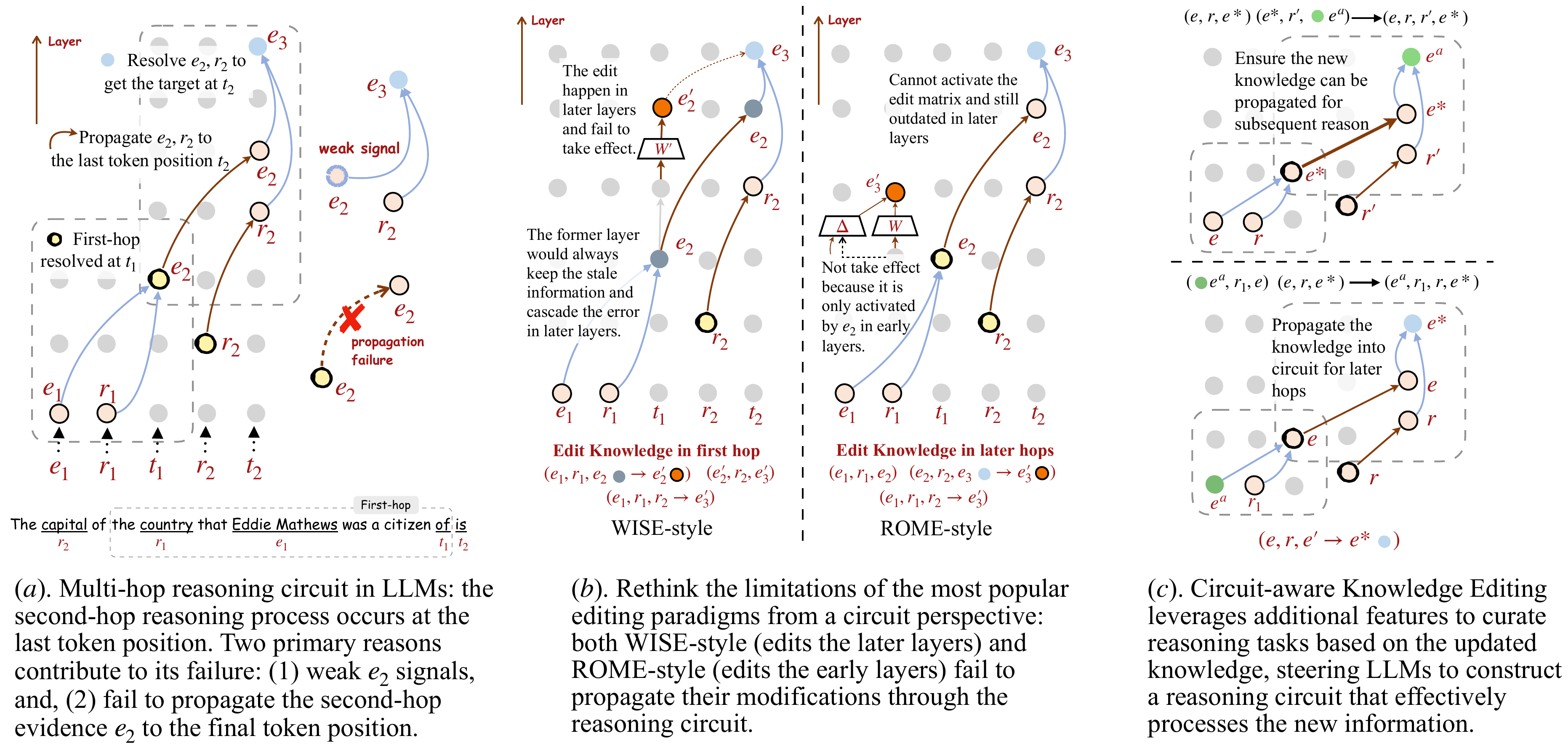}
\caption{An overview of our work.
}
\label{fig:circuit_view}
\end{figure*}
Our work specifically addresses the poor performance of edited models in downstream reasoning tasks that involve the updated knowledge~\cite{zhong2023mquake,zhang2024locate}.
Consider a representative case in Figure~\ref{fig:overview}
: after editing `\emph{Eddie Mathews, citizenship, United States → United Kingdom}', models correctly answer direct queries but fail multi-hop reasoning like `\emph{The capital of the country that Eddie Mathews was a citizen of is?}' (still outputting `\emph{Washington D.C.}'). 
Critically, this is not merely an editing artifact: vanilla LLMs often correctly answer single-hop questions while failing their multi-hop counterparts~\cite{yang2024largelanguagemodelsperform,biran-etal-2024-hopping}, suggesting deeper architectural limitations in knowledge utilization.

We trace these limitations to a misalignment between KE strategies and the inherent reasoning architectures of LLMs. 
To investigate this disconnect, we examine how LLMs leverage knowledge in downstream reasoning tasks.
Recent analysis suggests that knowledge is not merely statically stored but dynamically activated through specialized circuits~\cite{yao2024knowledge,biran-etal-2024-hopping,yu2025back}. 
However, these analyses overlook the phenomenon of LLM failures in reasoning circuits and fail to explore the underlying causes.
Our investigation (\S\ref{sec:utilization}) delves deeper into reasoning circuits, analyzing their structure and identifying the reasons behind failures in multi-hop reasoning.
Specifically, the multi-hop reasoning emerges from coordinated computing circuits: early layers handle the first hop, extracting the bridge entity at the end-token of the first hop.
This bridge entity, along with second-hop relation information, is then routed to the last token position in the middle layers. 
Subsequently, later layers utilize this information at the last token position to complete the reasoning process (Figure~\ref{fig:circuit_view} (a)).
We then analyze the entity and relation information at the last token position in failed multi-hop reasoning cases. 
Our observations reveal that critical information either fails to be properly routed to the last token position, or exhibits a weak signal, preventing effective reasoning.
This explains why current KE methods underperform (\S\ref{sec:edit}): they optimize for isolated parameter changes rather than circuit-level integration needed for compositional reasoning (Figure~\ref{fig:circuit_view}b). 

To bridge this fundamental gap, we propose \textbf{C}ircuit-\textbf{a}ware \textbf{K}nowledge \textbf{E}diting (CaKE) in \S\ref{sec:method}. 
Unlike methods that only update localized static knowledge, CaKE actively constructs reasoning circuits that enable dynamic application of edited knowledge in downstream tasks.
We first design circuit-aware training data that integrates across distinct segments of the reasoning process to force the LLM to leverage updated knowledge for latent reasoning (Figure~\ref{fig:circuit_view}c). 
Remarkably, we find that only a few such samples are sufficient to integrate knowledge across the reasoning circuit while maintaining strong general ability. 
Moreover, to prevent unintended data leakage, we construct these data using ad-hoc features~\cite{zhang2024cooccurrence} that are temporarily associated with the entities, such as \emph{`Japan is colored green. The capital city of the country colored in green is'}.
Finally, we guide LLMs to establish reasoning circuits by training with the curated data.
Extensive experiments (\S\ref{sec:exp}) demonstrate CaKE's effectiveness: it outperforms existing knowledge editing methods on the MQuAKE multi-hop benchmark for both LLAMA3-8B-Instruct~\cite{dubey2024llama} and Qwen2.5-7B-Instruct~\cite{yang2024qwen2}. Notably, \ours achieves this while being more memory-efficient than alternatives and successfully scales to larger models like LLAMA3-70B-Instruct.

\section{Analyzing Reasoning Circuit in LLM}
\label{sec:utilization}
\subsection{Data Preparation}
We employ the WikiData subset proposed by ~\citet{biran-etal-2024-hopping} and name it \emph{HoppingTooLate}, which contains 82,021 two-hop queries.
We denote each fact as a triplet $(e,r,e')$, where $e$ is the head entity, $r$ is the relation, and $e'$ is the tail entity.
We view two-hop queries as $(e_1, r_1, e_2)$ and $(e_2, r_2, e_3)$, where $e_1$ is the source entity, $e_2$ is the bridge entity, and $e_3$ is the target entity.
We focus on the latent reasoning framework to evaluate whether a model can output the expected answer $e_3$ directly given the composite query $(e_1,r_1,r_2,?)$.
For example, for the two facts \emph{(Eddie Mathews, country\_citizenship, United States)},\emph{(United States, capital, Washington D.C.)}, the composite query is \emph{(Eddie Mathews, country\_citizenship, capital,?)}.
We transform the question into the natural language expression: `\emph{The capital of the country that Eddie Mathews was a citizen of is?'.}
In addition, we follow \emph{HoppingTooLate} and define $t_1$ as the last token of the first-hop prompt (e.g., \emph{`the country that Eddie Mathews was a citizen \underline{of}'}) and $t_2$ as the last token of the whole two-hop prompt (e.g., `\emph{The capital of the country that Eddie Mathews was a citizen of \underline{is}')}.

\subsection{Multi-hop Reasoning Circuit}
Building on the insights from prior work~\cite{biran-etal-2024-hopping,yao2024knowledge}, we can define a structured circuit mechanism for multi-hop reasoning in transformer-based LLMs, as illustrated in Figure~\ref{fig:circuit_view}(a).
The  three distinct computational phases:
1) The model processes the initial relation $r_1$ and entity $e_1$, encoding the bridge entity $e_2$ in the final token position of the first prompt segment ($t_1$).
2) Critical features, including $e_2$ and the second relation, $r_2$ are transferred to the last token position $t_2$, preparing for final resolution.
3) The model computes the target $e_3$ by resolving $r_2$ and $e_2$, giving the result in the final token position.
Hence, based on the linearity theory~\cite{hernandezlinearity}, multi-hop reasoning in LLM can be formalized as:
\begin{equation}
    F_n(F_{n-1}(e_{n-1}, r_{n-1}), r_n)
\end{equation}
Each function $F_{n-1}$ produces a bridge entity $e_n$ for subsequent computation, demonstrating how intermediate results propagate vertically through network layers.

\begin{table}[t]
\centering
\resizebox{0.5\textwidth}{!}{
\begin{tabular}{cccccccc}
\toprule
\multirow{2}{*}{Model} & \multirow{2}{*}{Metric} & \multicolumn{2}{c}{Correct} & \multicolumn{2}{c}{Inconsistent} & \multicolumn{2}{c}{Incorrect} \\
& & Cases & Layer & Cases & Layer & Cases & Layer \\ \hline
\multirow{4}{*}{LLAMA3}& $e_2$ from $t_1$ & 63.1\% & 6.3 & 75.2\% & 6.0 & 48.7\% & 8.2  \\
 & $e_2$ from $t_2$ & 67.8\% & 13.2 & 59.8\% & 9.8 & 17.7\% & 21.1  \\
 & $r_2$ from $t_2$ & 66.9\% & 14.0 & 49.0\% & 13.8 & 28.1\% & 13.7  \\
 & $e_3$ from $t_2$ & 56.5\% & 18.8 & 22.7\% & 20.7 & 18.3\% & 18.0  \\ \midrule
\multirow{4}{*}{Qwen2.5} & $e_2$ from $t_1$ & 71.2\% & 4.3 & 74.1\% & 4.7 & 46.7\% & 5.1  \\
 & $e_2$ from $t_2$ & 52.9\% & 7.9 & 63.7\% & 9.5 & 18.9\% & 13.5  \\
 & $r_2$ from $t_2$ & 75.8\% & 8.1 & 75.2 \% & 10.4 & 44.8\% & 9.7  \\
 & $e_3$ from $t_2$ & 71.2\% & 16.4 & 39.4\% & 17.4 & 25.2\% & 11.4  \\ 
\bottomrule
\end{tabular}
}
\caption{The results of LLAMA3-8B-Instruct (32 layers) and Qwen2.5-7B-Instruct (28 layers). Cases are the percentage of data we can detect the information, and Layer is the mean of the earliest layer where the required information is detected.}
\label{tab:layer}
\end{table}
\subsection{Circuit in Failure Phenomena}
Then, we aim to understand why language models sometimes fail at multi-hop reasoning despite successfully answering individual single-hop questions. 
For instance, a model may correctly answer \emph{`the capital of Russia'} with \emph{`Moscow'} and \emph{`the country of citizenship of Fyodor Dostoyevsky'} with \emph{`Russia'}, yet fail to answer the multi-hop question \emph{`the capital of the country of citizenship of Fyodor Dostoyevsky is'} correctly. 
To systematically analyze this issue, we focus on the second hop of reasoning, as the model typically performs well on the first hop.
We categorize the data from the \emph{HoppingTooLate} dataset\footnote{We filter out short-cut cases as done by \citet{biran-etal-2024-hopping}.} into three subsets based on the model's behavior:
\textbf{Correct}: The model answers both single-hop questions $(e_1,r_1,e_2)$ and $(e_2,r_2,e_3)$ correctly, as well as the multi-hop question $(e_1,r_1,r_2,?)$.
\textbf{Inconsistent}: The model answers both single-hop questions correctly but fails on the multi-hop question. 
However, we observe that some questions in the \emph{Correct} set share the same 'bridge' entity $e_2$,  even though they originate from distinct subject-relation pairs, that the model answers correctly. $(e_1',r_1',r_2,?)$. 
This suggests that while the model can leverage knowledge in some contexts, it fails to generalize, indicating reasoning gaps rather than missing knowledge.
\textbf{Incorrect}: The model answers both single-hop questions correctly but fails on the multi-hop question in all contexts $(e_1',r_1',e_2)$. 
This implies a complete failure to employ the knowledge for multi-hop reasoning.
To investigate these failure modes, we check whether the models construct the reasoning circuit by monitoring key variables ($e_1$, $e_2$, and $r_2$) at critical positions ($t_1$ and $t_2$) across the model's layers using the PatchScope as \citet{biran-etal-2024-hopping} did.
Our analysis reveals several interesting patterns, extending beyond the `hopping too late' problem identified by \citet{biran-etal-2024-hopping}.

We list the results in Table~\ref{tab:layer}.
For the \emph{correct} subset, we observe strong evidence of the reasoning circuit functioning as expected:
a large portion of $e_2$ is detected at both $t_1$ ($e_2$ from $t_1$) and $t_2$ ($e_2$ from $t_2$) in both LLAMA3 and Qwen2.5 models.
The model correctly uses the $r_2$ and $e_2$ information at $t_2$ to produce the final answer $e_3$.
Contrastly, in the \emph{Incosistent} subsets, we can find that despite detecting $e_2$ and $r_2$ at $t_2$, the model often fails to produce the correct $e_3$ answer ($e_3$ from $t_2$: only 22.7\% in LLAMA3 and 39.4\% in Qwen2.5 of cases we can detect at $t_2$).
We hypothesize that the $e_2$ information, though present, may be insufficient to trigger the second-hop reasoning circuit, leading to the failure to execute the function $F(e_2, r_2)$ effectively.
What's more, in the \emph{Incorrect} subsets, we can find that the needed $e_2$ information is rarely detected at the $t_2$ position ($e_2$ from $t_2$: Only 17.7\% in LLAMA3 and 18.9\% in Qwen2.5).
Even when $e_2$ is detected, it typically emerges in much later layers (layer 21 in LLAMA3 and layer 13.5 in Qwen2.5), making it too late to be effectively utilized for the second-hop computation, aligned with \citet{biran-etal-2024-hopping}'s findings.
We conjecture the model fails to propagate $e_2$ to the $t_2$ position, resulting in the variable $e_2$ missing for conducting the $F(e_2,r_2)$ function. \looseness=-1
\begin{figure}
    \centering
    \includegraphics[width=\linewidth]{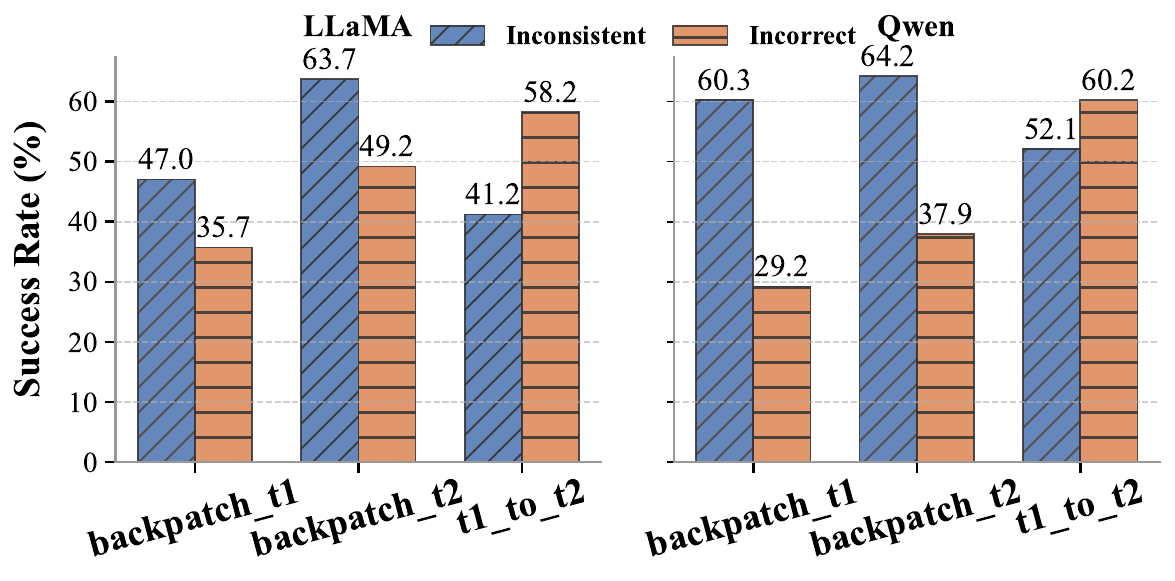}
    \caption{Results of the intervention on the failure cases in multi-hop reasoning of LLAMA3 and Qwen2.5.}
    \label{fig:enhance}
\end{figure}
\paragraph{Evaluation}
To test our hypothesis, we conduct interventions to enhance the information at the detected layers to see if we can improve the model's performance in these failure cases.
We test three ways: back-patching the $t_1$ and $t_2$ position as~\citet{biran-etal-2024-hopping} did, which would enhance the information at the position, and cross-position patching the information from $t_1$ to the $t_2$ position, which explicitly propagates the information from $t_1$ to $t_2$ (details in Figure~\ref{fig:patch_type} in Appendix).
From the results in Figure~\ref{fig:enhance}, we can find a high success rate for all the inconsistent and incorrect cases, but they demonstrate different paradigms.
For the inconsistent cases, back-patching would lead to better performance, while for the incorrect cases, patching knowledge from the $t_1$ to $t_2$ usually shows better outcomes. 
This proves our previous hypothesis that for the incorrect cases, due to the 
\textbf{propagation failure}, the model fails to move the $e_2$ to $t_2$ position, and manual routing via cross-patching can mitigate the issue.
Meanwhile, for inconsistent cases, amplification via back-patching compensates the \textbf{weak signal} when valid $e_2$ representations reach $t_2$ but lack sufficient magnitude for subsequent reasoning. \looseness=-1

\section{Circuits-aware Knowledge Editing}
Building on our previous reasoning analysis, we rethink the reason why current knowledge editing methods fail under multi-hop reasoning circumstances despite their great performance under single-fact editing. 
\subsection{Rethinking KE from the Circuit View}
\label{sec:edit}
Here, we aim to figure out what happens when we edit the model with the current KE methods.
\paragraph{Unified Editing Details}
When updating a piece of knowledge $(e, r, o \rightarrow o')$, the most popular knowledge editing techniques would modify the parameters that are responsible for the knowledge. 
There are two kinds of paradigms:
editing the Feed-Forward Networks (FFN) in the early layers, such as ROME~\cite{meng2022locating} and MEMIT~\cite{mengmass} or modifying the later layers' FFN output, like WISE~\cite{wang2024wise} and T-Patcher~\cite{tpatcher}.
This is mainly based on the key-value memory features of the FFN~\cite{geva2020transformer}.
However, some studies have queried the effectiveness of these localization settings~\cite{chang2024localization,hase2024does} as the localization area is not correlated to the performance of the knowledge editing methods.
Here, we propose a unified view of the mechanisms and limitations from the circuit perspective.
ROME-style would modify the weight $\boldsymbol{W}$ with a perturbation $\boldsymbol{\Delta}$ and obtain a new weight $\boldsymbol{W}'= \boldsymbol{W}+\boldsymbol{\Delta}$.
When calculating the $\Delta$, ROME-style methods, apply the \emph{least squares estimation} and \emph{null space constraint} to make sure the $\Delta$ is only activated by the corresponding entity representation $\boldsymbol{e}_{in}$ and keep the original output for other representations. 
In parallel, WISE-style editing methods would directly introduce the new weight $\boldsymbol{W}'$ that would be activated by the related representation $\boldsymbol{e}_{in}$, and $\boldsymbol{W}'$ would encode the updated knowledge.(More details in Appendix~\ref{app:method}).
Hence, these two editing paradigms can be represented uniformly by a gated function $\mathcal{G(\cdot)}$: 
\begin{equation}
    \text{FFN}_\text{out}(\mathbf{x}) = \underbrace{\boldsymbol{W}\mathbf{x}}_{\text{Original term}} + \mathcal{G}(\mathbf{x}) \cdot \underbrace{\boldsymbol{\delta}(\mathbf{x})}_{\text{Edit term}}
\end{equation}
\begin{equation}
    \mathcal{G}(\mathbf{x}) = \begin{cases} 
1, & \mathbf{x} \in \boldsymbol{e}_{in} \  \\
0, & \text{otherwise}
\end{cases}
\end{equation}
Here, for ROME-style method,$\boldsymbol{\delta}(\mathbf{x}) = \boldsymbol{\Delta}\mathbf{x}$ and for WISE-style method, $\boldsymbol{\delta}(\mathbf{x}) = (\boldsymbol{W}' - \boldsymbol{W})\mathbf{x}$.
When the gating function $\mathcal{G}(\cdot)$ is activated by the input $\mathbf{x}$, the edit term $\boldsymbol{\delta}(\mathbf{x})$ is applied, thereby modifying the knowledge within the computational circuit.

\paragraph{Defect from circuit view}

\begin{table*}[t]
\centering
\small
\renewcommand{\arraystretch}{1.2}
\resizebox{0.9\textwidth}{!}{
\begin{tabular}{cc|cc|cc|cc}
\toprule
\multirow{2}{*}{\textbf{Method}} & \multirow{2}{*}{\textbf{Model}}  & \multicolumn{2}{c|}{\textbf{MQUAKE-CF}} & \multicolumn{2}{c|}{\textbf{MQUAKE-CF-v2}} & \multicolumn{2}{c}{\textbf{MQUAKE-T}} \\
\cmidrule(lr){3-4} \cmidrule(lr){5-6} \cmidrule(lr){7-8}
&& \textbf{H-Acc.$\uparrow$} & \textbf{MAcc.$\uparrow$} & \textbf{H-Acc.$\uparrow$} & \textbf{MAcc.$\uparrow$} & \textbf{H-Acc.$\uparrow$} & \textbf{MAcc.$\uparrow$} \\
\midrule
Pre-edited & \multirow{9}{*}{\rotatebox{90}{{LLaMA3-8B-Ins}}}& 79.0  & 27.0  & 78.4& 28.6 & 71.0 & 5.3  \\
\midrule
AdaLoRA&  & 66.0 & \underline{27.6}  & 64.7  & 24.6 & \textbf{92.3}  & 66.0\\
WISE& & 38.2 & 24.0 & 37.2 & 21.0 & 63.5 & \underline{62.9}\\
MeLLo& & 16.5 & 16.1 & 19.5 & 16.0 & 42.3 & 50.1 \\
ROME& & \underline{86.8} & 17.6 & \underline{86.4}  & 15.5 & 89.5  & 8.4  \\
MEMIT& & 76.3  & 11.5  & 74.0 & 10.0 &86.0 & 3.7 \\
AlphaEdit& &  66.1  & 10.1  & 63.7 & 8.5  & 73.4 & 1.0 \\
IFMET $^{\clubsuit}$& & 81.9 & 23.2 & 75.3& \underline{36.5} & 82.1 & 46.1 \\
CaKE(ours) & &  \textbf{90.6}  & \textbf{57.3} & \textbf{90.1}  & \textbf{57.1} & \underline{91.5}  & \textbf{81.4}\\
\midrule
\midrule
Pre-edited & & 75.6 &  34.7  & 76.8 & 37.7  & 60.1 & 15.6\\
\midrule
LoRA & \multirow{3}{*}{\rotatebox{90}{{L-70B}}}  & \underline{93.1} & \underline{53.2} & \underline{90.5} & \underline{50.2} & \underline{90.1} & \underline{90.6} \\
MeLLo& & 8.0 & 6.4 & 8.6 & 9.9 & 11.6 & 32.9 \\
CaKE(ours) & & \textbf{93.5} & \textbf{65.4} &  \textbf{93.3} & \textbf{63.3} &\textbf{91.1} & \textbf{94.6} \\
\bottomrule
\end{tabular}
}
\caption{Comparison of \ours with existing methods on MQuAKE for LLAMA3-8B-Instruct and LLAMA3-70B-Instruct.  Due to the computational limitations, we just ran the LoRA and MeLLo in the 70B model. The best results are highlighted in bold, while the second-best results are underlined. ${\clubsuit}$ means the results are based on our re-implementation since the original code is not open by the authors, and we will update it after the source code is open.\looseness=-1}
\vspace{-8pt}
\label{tab:overall_comp}
\end{table*}

In single-hop knowledge editing, both these kinds of methods would give us the correct information, but for the multi-hop cases, they would fail.
As shown in Figure~\ref{fig:circuit_view} (b), both these layer-specific editing methods \textbf{cannot propagate the updated knowledge to the reasoning circuit}, leading to unsatisfactory multi-hop reasoning performances.
Consider the two-hop reasoning process from \S\ref{sec:utilization}: the model must first correctly compute $e_2 = F_1(\langle e_1, r_1 \rangle)$ in early layers. 
The representation of $e_2$ then propagates to the final token position $t_2$ (typically where answers are generated), where it combines with $r_2$ to compute $e_3 = F_2(\langle e_2, r_2 \rangle)$ in later layers.

WISE-style editing shows critical limitations when handling first-hop facts $(e_1,r_1,e_2 \rightarrow e_2')$ in multi-hop reasoning. 
As the edit is applied to later layers, the early layers remain unchanged and continue to produce the original $e_2$ representation during computation. 
This creates a fundamental mismatch: while the later layers perform the second-hop computation $F_2(\langle e_2, r_2 \rangle)$, they operate on the unmodified $e_2$ from early layers.
Consequently, the gating mechanism $\mathcal{G}(\cdot)$ designed for first-hop edits becomes effectively bypassed in the reasoning process.
Similarly, ROME-style editing fails when the edited fact $(e_2,r_2,e_3 \rightarrow e_3')$ serves as the second-hop question. 
For the edit to take effect, the gating function $\mathcal{G}(\cdot)$ must be activated by $e_2$ in early layers. 
However, $e_2$'s representation only appears after the first hop completes in the computational pathway - potentially after the edited layers. 
In this scenario, the gated function $\mathcal{G(\mathbf{x})}$ in earlier layers remains unactivated, causing the model to default to stale knowledge and produce incorrect answers. \looseness=-1
\subsection{Proposed Method: CaKE}
\label{sec:method}
Inspired by previous analysis, we propose a novel method, \textbf{C}ircuit-\textbf{a}ware \textbf{K}nowledge \textbf{E}diting (\textbf{CaKE}), which makes sure the models build the reasoning circuit with the updated knowledge.
As we show in the previous section, a successful reasoning circuit is one that, after editing the model’s knowledge, ensures: 
The updated computation
$F_1$  or  $F_2$ accurately reflects the new knowledge, and the bridge entity $e_2$  is correctly computed and propagated to  $t_2$.
Hence, simply editing a single layer or several layers is not enough to enable the circuit for reasoning.
Here, \ours comprises two key components: (1) generating circuit-aware data that explicitly requires reasoning with the updated knowledge, and (2) training the model to construct robust reasoning circuits that integrate the new knowledge.
\paragraph{Data Generation} 
For each updated knowledge item, we construct the following contexts to mitigate these issues:
(1) \textbf{Original Narrative}: We begin by generating straightforward factual statements that explicitly convey the updated information.
For example, when updating the fact  $k$: (PersonX, citizen\_country, Switzerland → Japan), we use the narrative representation: \emph{`PersonX is a citizen of Japan'} and generate several paraphrases.
These statements serve as the foundation for the model to learn the updated knowledge.
(2) \textbf{Circuit-aware Tasks}: Next, we design specialized reasoning scenarios that address two critical challenges: preventing \emph{failure propagation} and mitigating \emph{weak signals}, while ensuring that updated knowledge is properly integrated across different layers (in Figure~\ref{fig:circuit_view}c). 
Moreover, to avoid introducing extraneous knowledge that could leak into downstream evaluations—and to test the generalization of our method (inspired by prior research~\cite{zhang2024cooccurrence})—we incorporate ad-hoc features into these scenarios. 
Particularly, these tasks link the facts with intermediate attributes or reasoning steps and fall into two categories:
\textbf{Late-layer Knowledge Integration:} These tasks ensure that the updated knowledge is effectively learned in the later layers, alleviating issues such as \emph{weak signals and the limitations of ROME-style editing}. 
Take the fact $k$: (PersonX, citizen\_country, Switzerland → Japan) as an example; we construct a seed prompt like:
\emph{`Suppose \{random\_entity\_1\} wears red clothes,
\{random\_entity\_2\} wears blue clothes,
and \{PersonX\} wears green clothes.
The country of citizenship of the person in green is:'}
Here, the model is expected to output `Japan,' requiring it to employ the new fact $k$ in later layers.
\textbf{Reasoning Circuit Enhancement:} 
These tasks require the model to use the updated knowledge for subsequent reasoning, thereby mitigating \emph{propagation failure} and \emph{WISE-style's limitations}.
Following the same fact $k$, the seed prompt is
\emph{`In a book about countries, Japan is mentioned on page 6 of the book,
while China is mentioned on page 72.
On which page of the book is the country of citizenship of the \{PersonX\} shown?'}
Here, the model must first recall the updated citizenship (Japan) and then use this information to determine the correct page number (6).

For each relation type, we design these seed task templates and employ GLM-4-plus~\cite{glm2024chatglm} to generate diverse expressions following these templates (see Appendix~\ref{app:data_gen} for details). 
Specifically, we create 3 distinct samples per category for each edited fact, which our experiments show are sufficient to enable effective reasoning with the updated knowledge. 
This minimal data requirement demonstrates the efficiency of our approach in adapting models to new information.
\paragraph{Edit Training}
After obtaining the curated circuit-aware data $\mathcal{D}$, we fine-tune the LLM using LoRA, enabling the model to optimize its internal knowledge organization.
We minimize the cross-entropy loss $\mathcal{L}$ between the model's outputs and the ground-truth tokens expressing the updated fact: \looseness=-1
\begin{equation}
   \mathcal{L} = \mathbb{E}_{(\mathbf{x}, \mathbf{y}) \in \mathcal{D}}\left[-\sum_{t=1}^{|\mathbf{y}|} \log p(y_t \mid \mathbf{x}, \theta_{\text{LoRA}})\right]
\end{equation}
where $\theta_{\text{LoRA}}$ represents the LoRA parameters, $\mathbf{x}$ is the input prompt, and $\mathbf{y}$ is the desired updated output sequence.

\section{Experiments}
\label{sec:exp}

\begin{table*}[t]
\centering
\small
\renewcommand{\arraystretch}{1.2}
\resizebox{0.85\textwidth}{!}{
\begin{tabular}{cc|cc|cc|cc}
\toprule
\multirow{2}{*}{\textbf{Method}} & \multirow{2}{*}{\textbf{Model}}  & \multicolumn{2}{c|}{\textbf{MQUAKE-CF}} & \multicolumn{2}{c|}{\textbf{MQUAKE-CF-v2}} & \multicolumn{2}{c}{\textbf{MQUAKE-T}} \\
\cmidrule(lr){3-4} \cmidrule(lr){5-6} \cmidrule(lr){7-8}
&& \textbf{H-Acc.$\uparrow$} & \textbf{MAcc.$\uparrow$} & \textbf{H-Acc.$\uparrow$} & \textbf{MAcc.$\uparrow$} & \textbf{Hop-wise.$\uparrow$} & \textbf{MAcc.$\uparrow$} \\
\midrule
Pre-edited & \multirow{9}{*}{\rotatebox{90}{{Qwen2.5-7B-Ins}}}& 73.4   & 40.7  & 72.8 & 39.5  & 56.1 & 15.6\\
\midrule
AdaLoRA&   & 35.1 & 24.9 & 36.5  & \underline{25.9}  & 25.0 & 28.6\\
WISE& & 41.2 & 9.8 & 26.5 & 8.0 & 50.2  & 36.5\\
MeLLo& & 35.5 & 7.8 & 34.5 & 7.6 & 52.7 & \underline{56.5} \\
ROME& & 75.4  & 10.7   & 73.4 & 8.8 & 86.7 & 17.7  \\
MEMIT&   & 82.6  & 11.1 & 83.4  & 9.6  &  88.9 & 18.5 \\
AlphaEdit& & 73.8 & 12.6  & 75.1 & 10.5 &  82.2 & 17.2  \\
IFMET $^{\clubsuit}$ & & \underline{83.7} & \underline{25.7}  & \underline{84.6} & 24.5& \underline{90.0} & 52.8 \\
CaKE(ours) &  & \textbf{90.6} & \textbf{61.4} & \textbf{90.3} & \textbf{63.05} & \textbf{95.5} &  \textbf{87.8} \\
\bottomrule
\end{tabular}
}
\caption{Comparison of \ours with existing methods on MQuAKE on Qwen2.5-7B-Instruct.  The best results are highlighted in bold, while the second-best results are underlined. ${\clubsuit}$ means the results are based on our own implementation since the original code is not open by the authors, and we will update it after the source code is open.\looseness=-1}
\label{tab:qwen_overall_comp}
\end{table*}
\subsection{Experiment Settings}
We mainly utilize the multi-hop reasoning knowledge editing dataset MQuAKE~\cite{zhong2023mquake}, which considers different numbers of hops (from 2 to 4) and different positions of the knowledge used in the multi-hop questions.
We utilize three versions of the datasets: MQuAKE-CF-3k and MQuAKE-CF-3k-v2, which are two subsets that contain different question types and editing hopping numbers, and MQuAKE-T is a time-aware knowledge editing benchmark.
\paragraph{Baselines and Models}
We consider several knowledge editing baselines, including: IFMET~\cite{zhang2024locate}, AlphaEdit~\cite{fang2024alphaedit}, ROME~\cite{meng2022locating}, MEMIT~\cite{mengmass},WISE~\cite{wang2024wise} and MeLLo~\cite{zhong2023mquake}.
Here, AlphaEdit, ROME, and MEMIT are methods that edit the model's parameters at early layers; WISE adds additional parameters at later layers, and IFMET edits both the early and later layers' FFN to achieve better multi-hop reasoning performance.
MeLLo is a prompt-based retrieval-augmented method that keeps the model's parameters unchanged.
We conduct experiments on LLAMA-3-8B-Instruct, Qwen-2.5-7B-Instruct, and LLAMA-3-70B-Instruct.
\paragraph{Evalutation Metric}
Following \citet{zhong2023mquake}, we evaluate model performance using Multi-hop Accuracy (MAcc) and Hop-wise Answering Accuracy (H-Acc). 
MAcc measures the accuracy of multi-hop question answering, while H-Acc assesses correctness at each reasoning step. 
For both metrics, we consider a prediction correct if the ground-truth answer appears in the generated text as \citet{cohen2024evaluating,zhong2023mquake} did. Higher values indicate better reasoning capability.
For KE, we also need to consider locality, which ensures edits do not affect unrelated knowledge and abilities. To assess this, we evaluate the model on general benchmarks, including CommonsenseQA~\cite{talmor2019commonsenseqa}, BigBenchHard~\cite{suzgun2023challenging}, MMLU~\cite{mmlu}, and GSM8k~\cite{cobbe2021training}.

\subsection{Experiments Results}
\paragraph{Main Results} 
We show the results for LLAMA3-8B-Instruct in Table~\ref{tab:overall_comp} and Qwen2.5-7B-Instruct in Table~\ref{tab:qwen_overall_comp}. 
From the table, we can find that although current KE methods achieve high hop-wise accuracy (H-Acc.), their performance on the three versions of MQuAKE is quite low (with an average accuracy of less than 20\%). 
For example, MEMIT and ROME achieve over 80\% accuracy on single-hop questions in MQuAKE-v2; however, their accuracy on multi-hop reasoning drops to only around 10\%, indicating that the LLM fails to effectively utilize the updated knowledge during reasoning.
In contrast, \ours demonstrates significant improvements in multi-hop reasoning. 
In the LLAMA3-8B-Instruct model, \ours achieves accuracies of 57.3, 57.2 and 81.5 in MQuAKE-CF, MQuAKE-CF-v2 and MQuAKE-T, respectively, outperforming all the compared methods. 
Additionally, IFMET, which also considers different layers for multi-hop reasoning but neglects the information flow within the circuit, performs not as well as CaKE.
Moreover, when compared with RAG-based methods such as MeLLo, \ours also yields better results.
Furthermore, compared to the baseline LoRA tuning methods that simply incorporate the raw knowledge, the improvements observed with \ours underscore the effectiveness of our approach. 
Results in Qwen-2.5-Instruct also demonstrate the same phenomenon.

\begin{figure}
    \centering
    \includegraphics[width=\linewidth]{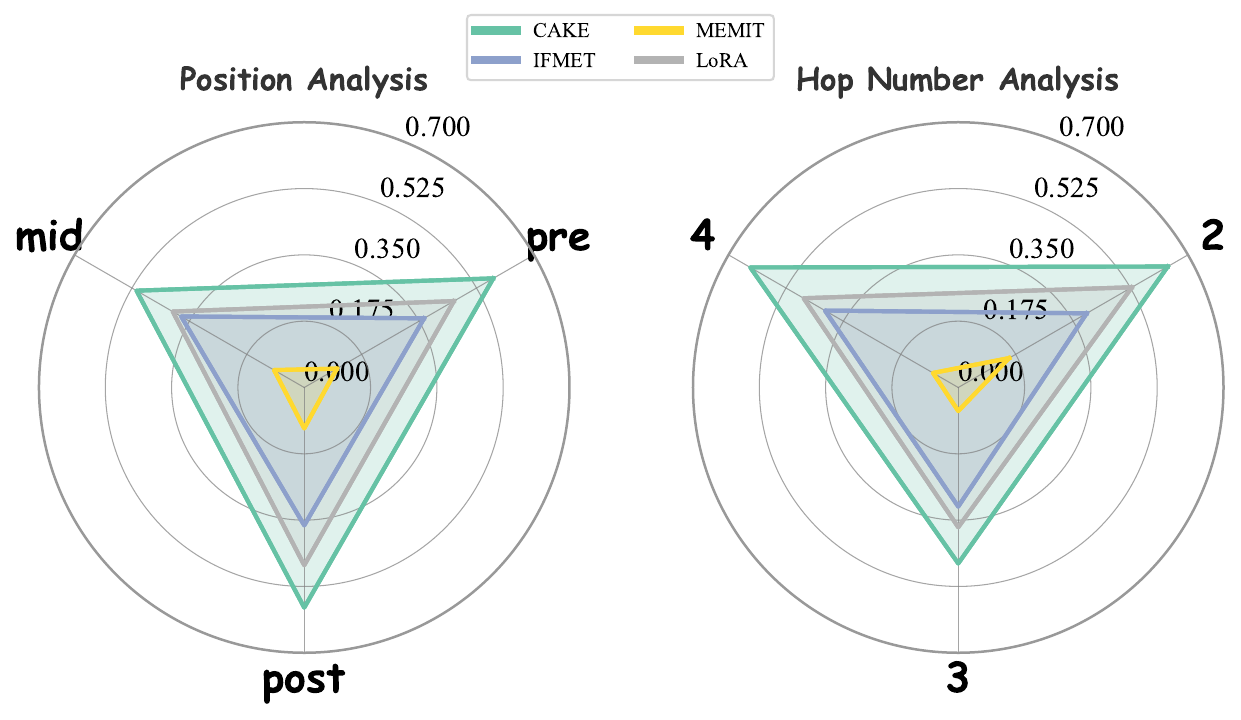}
    \caption{Accuracies of different number hops and edit-positions in MQuAKE-CF-3k-v2 on LLAMA3-8B-Instruct.}
    \label{fig:edit_position}
\end{figure}

\paragraph{Position and Number of Hop} We also compare the performance on different hops and positions in Figure~\ref{fig:edit_position}. 
Even when the model is trained solely on two-hop questions, CaKE yields improvements across varying numbers of editing hops. 
The benefits are particularly pronounced for four-hop questions, where methods like IFMET (designed only for two-hop scenarios) struggle. 
Besides, CaKE enhances performance regardless of the position of
the edited knowledge within the multi-hop questions, demonstrating the generalizability of CaKE.
\paragraph{Efficiency and Scalability} 
We evaluate computational efficiency in Table~\ref{tab:efficient}, demonstrating that \ours achieves better performance while requiring less memory than MEMIT with comparable editing times.
This efficiency advantage enables \ours to scale effectively to larger models, where it maintains superior performance on LLAMA3-70B-Instruct.
%
\begin{table}[t]
\centering
\resizebox{0.45\textwidth}{!}{
\begin{tabular}{ccccc}
\toprule
 & \textbf{CSQA} & \textbf{BBH} & \textbf{MMLU} & \textbf{GSM8k}  \\ \midrule
\textit{LLaMA3-8B-Ins} & 76.09 & 67.89 & 63.83 & 75.20 \\ \midrule
MEMIT & 76.08 & 67.88 & 63.82 & 75.21  \\ 
ROME & 72.98 & 61.37 & 62.95 & 74.59   \\ 
CAKE & 75.10 & 67.20 & 62.98 & 76.04 \\ 
\midrule
\midrule
\textit{Qwen2.5-7B-Ins} & 82.31 & 33.39 & 71.80 & 82.26 \\ 
\midrule
MEMIT & 82.39 & 37.37 & 71.80 & 81.96 \\ 
ROME & 72.57 & 34.22 & 63.38 & 72.21    \\ 
CAKE & 82.64 & 37.44 & 71.76 & 82.79  \\ \bottomrule
\end{tabular}
}
\caption{\textbf{Locality Performance} on several general benchmarks of \ours and other editing methods.}
\vspace{-8pt}
\label{tab:locality}
\end{table}
\section{Analysis}
\subsection{Locality Performance}
In this section, we evaluate the model's performance on general ability benchmarks to ensure that acquiring new knowledge does not compromise its overall capabilities. 
As shown in Table~\ref{tab:locality}, \ours achieves performance comparable to the original model on both the LLAMA3-8B and Qwen2.5-7B models across different kinds of tasks, including math, commonsense, and diverse understanding tasks.
\subsection{Case Analysis}
In this part, we show the cases in which the \textbf{CaKE helps the model learn the multi-hop reasoning circuit and other methods fail}.
For illustration, we consider the two-hop question: `The capital city of the country that Eddie Mathews was a citizen of is'. 
Here, the editing case is \emph{(Eddie Mathews, citizenship, United States $\rightarrow$ United Kingdom)}, and the updated model is expected to output `London'.
However, \ours gives the correct answer, while other methods fail: 
MEMIT gives us the `Moscow', AlphaEdit gives us `Birmingham', and LoRA gives us `not known'.
To further understand these differences, we analyze the computing circuit of each method to determine \emph{whether the updated model successfully propagates the bridge entity $e_2$ and relation $r_2$ to the last token $t_2$ position.}

\begin{figure}
    \centering
    \includegraphics[width=\linewidth]{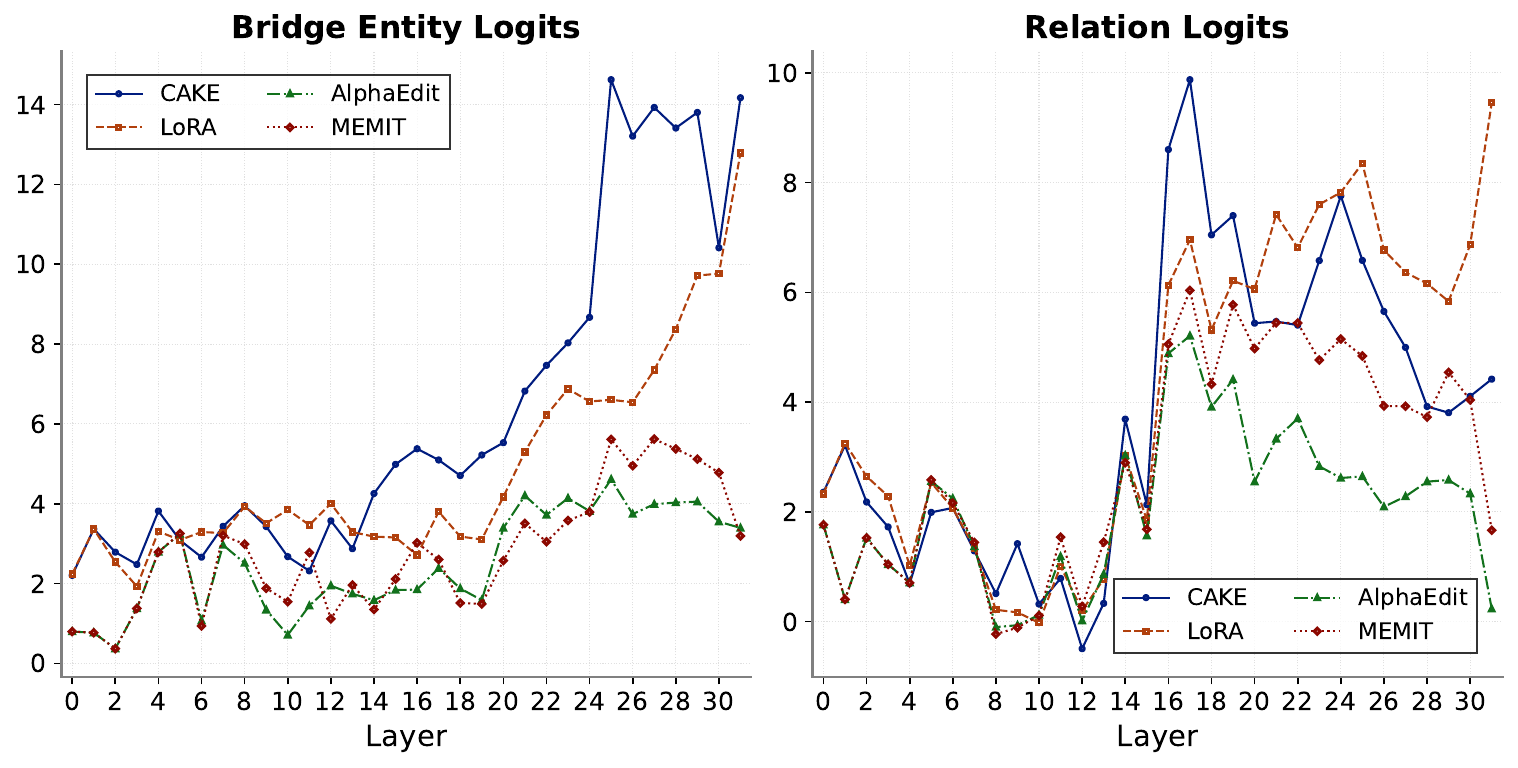}
    \caption{$e_2$ and $r_2$'s logits at $t_2$ in models after different knowledge editing methods.}
    \label{fig:case}
\end{figure}

Figure~\ref{fig:case} compares the logits of $e_2$ and $r_2$ at position $t_2$ across different editing methods.
Here, \ours generates significantly stronger logits for the bridge entity $e_2$ compared to AlphaEdit and MEMIT.
This demonstrates \ours's ability to propagate critical information to target positions for subsequent reasoning steps.
Similarly, \ours produces more prominent $r_2$ logits, indicating more robust circuit construction and information flow compared to baseline methods.
 
\section{Related Work}
\paragraph{Knowledge Learning and Editing}
Knowledge editing \cite{lampinen2025generalization,DBLP:conf/acl/JiangWWZZGLJSTL24,DBLP:conf/acl/SunLWMX0024,DBLP:conf/emnlp/HsuehHLLFHC24,DBLP:conf/acl/PowellGH24,DBLP:conf/emnlp/000400CZ024,DBLP:conf/emnlp/RoznerBWL24,DBLP:conf/emnlp/ZhangY0RWC24,DBLP:journals/corr/abs-2401-10471,DBLP:conf/cikm/ShiTWZZL24,huang2024reasons,guo2024mechanistic,DBLP:journals/csur/WangZLZCL25,feng2025geoedit,yang2025mirage,li2024understanding,huang2024can} has emerged as a promising approach for updating models in an ever-changing world. Current knowledge editing methods typically follow one of several strategies: modifying the MLP components in earlier layers \cite{meng2022locating, mengmass}, enhancing the MLP in later layers \cite{grace}, or retrieving relevant facts as prompts \cite{jiang-etal-2024-learning, zhong2023mquake}. However, most existing knowledge editing techniques concentrate on simple factual updates and frequently fail to generalize to more complex downstream tasks, such as multi-hop reasoning scenarios. 

\paragraph{Model Interpretability}
Knowledge editing is primarily based on the intrinsic knowledge mechanisms of neural models’ “black boxes” \cite{ferrando2024primer}. 
Consequently, understanding how knowledge in LLMs is acquired and stored has garnered significant attention \cite{wang2024knowledge}.
Recent studies \cite{DBLP:conf/nips/ZhouLX0SMMEYYZG23} demonstrate that most knowledge is learned during the pretraining stage and is predominantly stored in the Feed-Forward Network \cite{geva2020transformer}. 
Beyond these localized findings, researchers \cite{geva2023dissecting,yao2024knowledge} have investigated the computational circuits—the pathways connecting Transformer components—to elucidate how LLMs perform knowledge recall. 
Building on this, subsequent work has explored the relationship between knowledge editing and these circuits \cite{ge2024circuits}. 
In contrast, our work focuses on the mechanisms underlying multi-hop reasoning in LLMs and aims to improve the generalization of edited knowledge. \looseness=-1
\section{Conclusion}
In this paper, we identify that existing knowledge editing methods fall short due to their isolated parameter adjustments by examining the multi-hop reasoning circuits within LLMs.
We present \ours, a method designed to align knowledge editing with the inherent reasoning architectures of LLMs. 
\ours incorporates circuit-aware tasks that compel the model to dynamically integrate and utilize new knowledge during reasoning.
Experimental results demonstrate that CaKE achieves generalizable multi-hop knowledge editing.


\section*{Limitation}
\paragraph{Dataset}
Our work primarily focuses on the factual knowledge embedded in large language models (LLMs) and their capacity for multi-hop reasoning over these facts. 
We recognize that LLM reasoning also encompasses other domains—such as long-form mathematics and reverse-curse reasoning—that merit further investigation.
\paragraph{Reasoning Pattern}
As discussed in the previous analysis, we concentrate on direct reasoning phenomena. 
Current LLMs have shown impressive capabilities in slow-thinking paradigms, including chain-of-thought and reflective reasoning. 
Beyond direct reasoning, enhancing the utilization of knowledge within these paradigms represents an important avenue for future research.
\paragraph{Fine-grained Circuit Components}
Our analysis revealed relational information within the circuits; however, \ours currently does not delve deeply into these relationships. 
We believe that a more focused investigation into these components is necessary. Additionally, while our study emphasizes general circuit behavior, developing a more concise and effective method for knowledge editing remains an exciting challenge for future work.
\paragraph{Data Attribution}
Although we demonstrate the ability to construct reasoning circuits using curated data, the connection between a model's acquired abilities in its parameters and its training data is still underexplored. A deeper understanding of this relationship could lead to more efficient training processes and the generation of higher-quality synthetic data.

\section*{Acknowledgements}
We thank Lucas Bandarkar and other members of the UCLA PlusLab \& NLP group, as well as anonymous reviewers, for their constructive feedback and discussions.

This work was supported by the National Natural Science Foundation of China (No. 62576307, No. NSFCU23B2055, No. NSFCU19B2027), the Fundamental Research Funds for the Central Universities (226-2023-00138), Yongjiang Talent Introduction Programme (2021A-156-G), Tencent AI Lab Rhino-Bird Focused Research Program (RBFR2024003), Ningbo Natural Science Foundation (2024J020), Information Technology Center and State Key Lab of CAD\&CG, Zhejiang University.


\bibliography{example_paper}

\newpage
\appendix
\section*{Appendix}
\section{Setting Detail}
\paragraph{Dataset}
We list the details of the dataset in Table~\ref{tab:dataset}.
\begin{table}[ht]
\centering
\resizebox{0.5\textwidth}{!}{
\begin{tabular}{crrr}
\toprule
Model & Correct & Inconsistent &  Incorrect \\ \midrule
LLaMA3-8B-Ins. & 1,005 & 1,032 & 1,240 \\ 
Qwen2.5-7B-Ins. & 241 & 252 & 275 \\ \bottomrule
\end{tabular}
}
\caption{The dataset we used in the analysis.}
\label{tab:dataset}
\end{table}

\paragraph{Environment Setting}
We run our experiments on 2 NVIDIA-A800 GPUs.
For data generation, we utilize glm-4-plus and glm-4-air and a total of 10,000,000 tokens (about 20 dollars) to generate all synthetic data for the whole 7,867 data samples.
The cost is approximately 0.002 dollars per edit, which also demonstrates the efficiency of \ours method.
We use LLM-Eval~\cite{eval-harness} to test the model's general performance.
\paragraph{Data Generation}
\label{app:data_gen}
We first construct the question template $\mathcal{T}$ for each relation type, and we list some of them in Table~\ref{tab:template}.
We then generate the data using the following prompt:
\begin{tcolorbox}[mybox, title={Prompt for Constructing the circuit-aware data}]
Here are some question templates for the specific relation.
As you can see, the question use the knowledge in the input to conduct reasoning in different hops for multi-hop reasoning. Please generate 3 different questions that share the same features as the template. Please return a python json file.
\{$\mathcal{T}$\} Here is the input question:
\end{tcolorbox}
It should be noted that we do not ask the model to strictly follow the expression of the template, and we also show some data samples in Appendix~\ref{app:data_gen} to show the diversity of the generated data.

\begin{table*}[ht]
\centering
\resizebox{\textwidth}{!}{
\begin{tabular}{lll}
\toprule
\textbf{Knowledge Type} & \textbf{Template} & \textbf{Answer} \\
\midrule
\multirow{4}{*}{\makecell[l]{\{target\_person\} works \\in the field of \{target\_field\}}}. & \makecell[l]{In a book related to different fields, Section A discusses \{random\_field\},\\ Section B discusses \{random\_field\}, and Section C discusses \{target\_field\}.\\ If you want to learn about \{target\_person\}'s field, \\which section should you read? } & \makecell[l]{The working field of \{target\_person\} \\is discussed in Section C. }\\
\cline{2-3}
 & \makecell[l]{In a biography book, Section A discusses the life of \{random\_person\}, \\ Section B discusses the life of \{random\_person\},\\ and Section C discusses the life of \{target\_person\}. \\The field of the person in Section C is?} & \makecell[l]{The person in Section C \\works in the field of \{target\_field\}. }\\
\hline
\multirow{4}{*}{\makecell[l]{\{target\_person\} speaks \\ the language of \{target\_language\}.}}  & \makecell[l]{The following facts are known: 1. \{target\_person\} wears red clothes. \\ 2. \{random\_person\} wears blue clothes. \\3. \{random\_person\} wears green clothes.\\The language that the person in red clothes speaks is?} & \makecell[l]{The language that the person in red clothes \\speaks is \{target\_language\}. }\\
\cline{2-3}
 & \makecell[l]{At a global company: \\
 \{target\_language\}-speaking employees work in Team A.\\ \{random\_language\}-speaking employees work in Team B. \\In which team would \{target\_person\} work when he/she is at work?} & \makecell[l]{\{target\_person\} would work in\\ Team A when he/she is at work.} \\
\bottomrule
\end{tabular}
}
\caption{Sample templates for generating the circuit-aware data.}
\label{tab:template}
\end{table*}

\section{Implementation Detail}
\subsection{Analyzing Method}
\label{app:analysis_tool}
\begin{figure}
    \centering
    \includegraphics[width=0.7\linewidth]{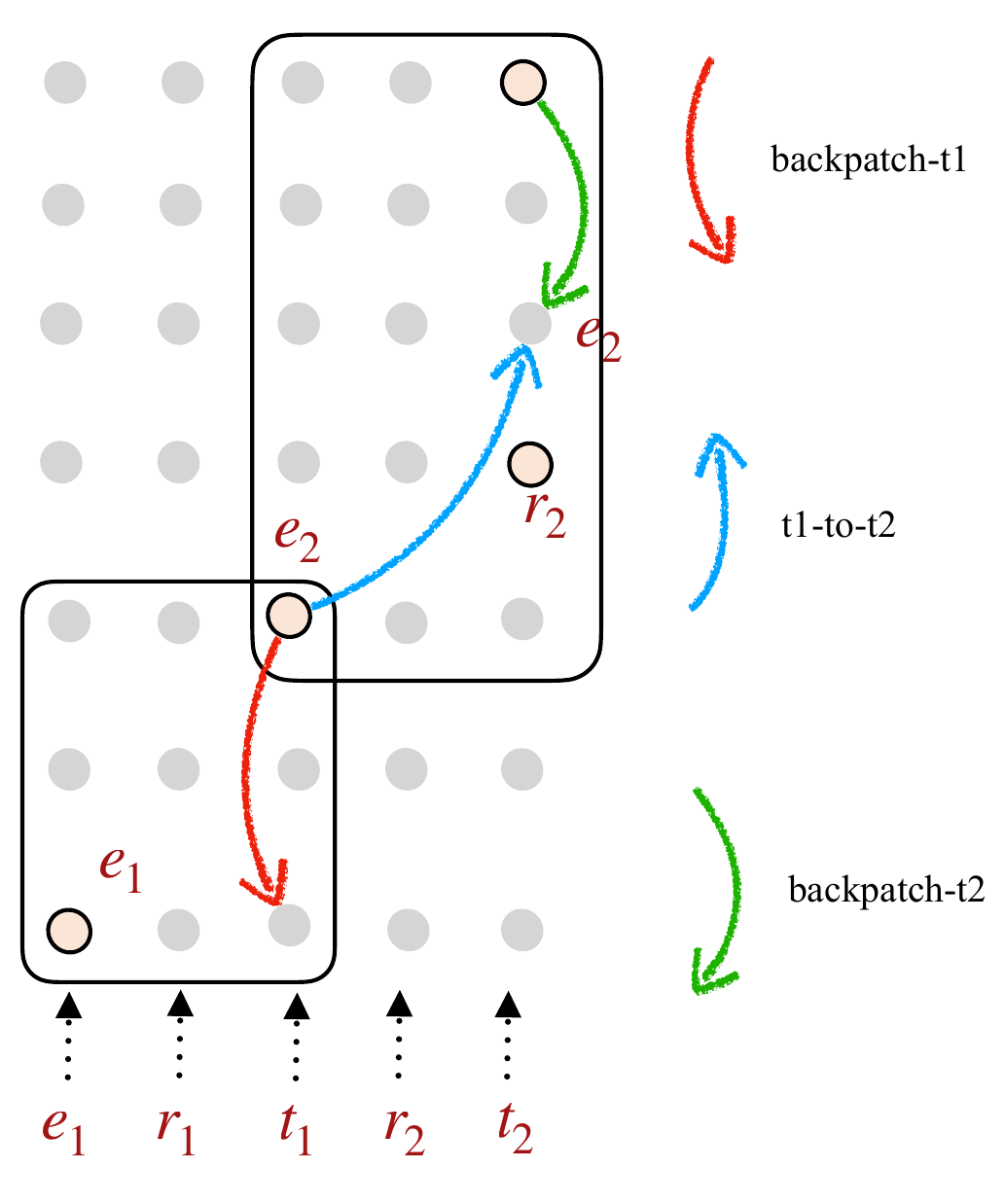}
    \caption{The way we conduct the backpatch and $e_1$ to $e_2$. We substitute the hidden representations from the source position to the target position.}
    \label{fig:patch_type}
\end{figure}

\paragraph{Patch Scope} 
The process is carried out as follows. 
First, a source prompt, a source token, and a source layer are provided. The prompt is processed through the model’s forward computation, and the hidden representation \( v \) of the source token at the specified layer is extracted and stored. This representation \( v \) is the focus of our investigation, as we seek to determine whether it encodes a specific entity.  Next, we employ the same prompt used by \citet{ghandehariounpatchscopes}:  
``Syria: Syria is a country in the Middle East. Leonardo DiCaprio: Leonardo DiCaprio is an American actor. Samsung: Samsung is a South Korean multinational corporation. x''
This prompt is passed through the model, but the hidden representation of `x' is replaced with \( v \) at a chosen target layer. The forward computation then proceeds, and the resulting generated text is analyzed to evaluate the effects of this substitution.
We conduct different patch analyses and show them in  Figure~\ref{fig:patch_type}.
When we conduct back-patch and cross-patch, the source prompt and target prompt are the same.

\subsection{Editing Method}
\label{app:method}
We utilize EasyEdit~\cite{easyedit} to conduct our editing experiments. 
For ROME, MEMIT, WISE, AlphaEdit, and MeLLo, we directly employ the original parameters provided by their respective papers.
Below, we introduce these methods in detail and describe our implementation.
\paragraph{ROME and MEMIT}
ROME leverages causal analysis to identify knowledge within specific MLP layers and modifies the corresponding weight matrix using least squares approximation. 
It operates under the strong assumption that the MLP layers primarily store knowledge and injects new information into these layers iteratively using a Lagrangian remainder.
In our experiments, we edit the 5th layer of both LLAMA3-8B-Instruct and Qwen2.5-7B-Instruct.

Similarly, MEMIT assumes that the FFN layers function as a knowledge key-value store. 
It directly modifies the parameters of selected layers through least squares approximation. 
Unlike ROME, which updates a single layer, MEMIT is a multi-layer editing algorithm capable of simultaneously updating hundreds or thousands of facts
\paragraph{IFMET}
IFMET builds upon MEMIT by not only modifying earlier MLP layers in transformers but also adjusting later layers to enhance multi-hop reasoning for the edited knowledge.
To ensure the updated knowledge propagates effectively, IFMET constructs an additional support set that reinforces learning in later layers.
Based on our analysis in \S\ref{sec:utilization}, we edit layers [17,18,19,20] for LLAMA3-8B-Instruct and layers [15,16,17,18] for Qwen2.5-7B-Instruct.
\paragraph{WISE}
WISE represents a different approach to model editing, focusing on later layers instead of earlier ones. 
It modifies the model's FFN output using a gating mechanism:
\begin{equation}\label{equ:ffn_out}
    \text{FFN}_\text{out}(\mathbf{x}) = \begin{cases}
    \mathcal{G}(\mathbf{x}) \cdot \mathbf{W}_{v'} &\mbox{if $\mathcal{G}(\mathbf{x})  > \epsilon$,}\\
    \mathcal{G}(\mathbf{x}) \cdot \mathbf{W}_{v} &\mbox{otherwise.}
    \end{cases}
\end{equation}
Here, $\mathcal{G}(\mathbf{x})$ is a gate function that computes the activation score of the hidden reprsentation: $\Vert \mathcal{A}(\mathbf{x}) \cdot (\mathbf{W}_{v'} - \mathbf{W}_v) \Vert_2$.
If the gate is activated, the model uses the updated knowledge to generate responses; otherwise, it relies on the original knowledge. 
Different methods define the gate function differently, but the core idea is to ensure that the updated memory aligns with relevant question representations.

\begin{table}[ht]
\centering
\resizebox{0.5\textwidth}{!}{
\begin{tabular}{lcccc}
\toprule
\multirow{2}{*}{\textbf{Edit Method}} & \multicolumn{2}{c}{\textbf{LLAMA3-8B}} & \multicolumn{2}{c}{\textbf{Qwen2.5-7B}} \\
\cmidrule(lr){2-3} \cmidrule(lr){4-5}
& First\_hop & Second\_hop & First\_hop & Second\_hop \\
\midrule
ROME & \textbf{16.66} & 7.81 & \textbf{10.57} & 8.33 \\
WISE & 49.85 & \textbf{67.36} & 8.33 & \textbf{33.59} \\
\bottomrule
\end{tabular}
}
\caption{Performance comparison of edit methods across different positions for the edited fact.}
\label{tab:performance}
\end{table}

\paragraph{MeLLo}
MeLLo is a non-parametric editing method that modifies a model's knowledge through prompting rather than weight updates.
It maintains a memory of newly introduced facts and guides the model to decompose multi-hop queries into sub-questions. 
At each step, the model checks this memory to verify whether its existing knowledge contradicts the new facts.
We follow the prompt structure provided in the original MeLLo method. 
However, in our experiments, we observe that the model struggles to consistently adhere to the intended reasoning pattern.

\paragraph{CaKE}
We utilize the original LoRA~\cite{hu2022lora} and add parameters in the FFN module in the model.
The hyperparameters are as follows:
\begin{itemize}[itemsep=1pt,topsep=0pt,parsep=0pt]
    \item epoch: [40, 50, 60]
    \item batch size: [4]
    \item learning rate: [1e-4]
    \item rank: [8]
    \item lora\_alpha: [32]
\end{itemize}

\begin{figure}
    \centering
    \includegraphics[width=\linewidth]{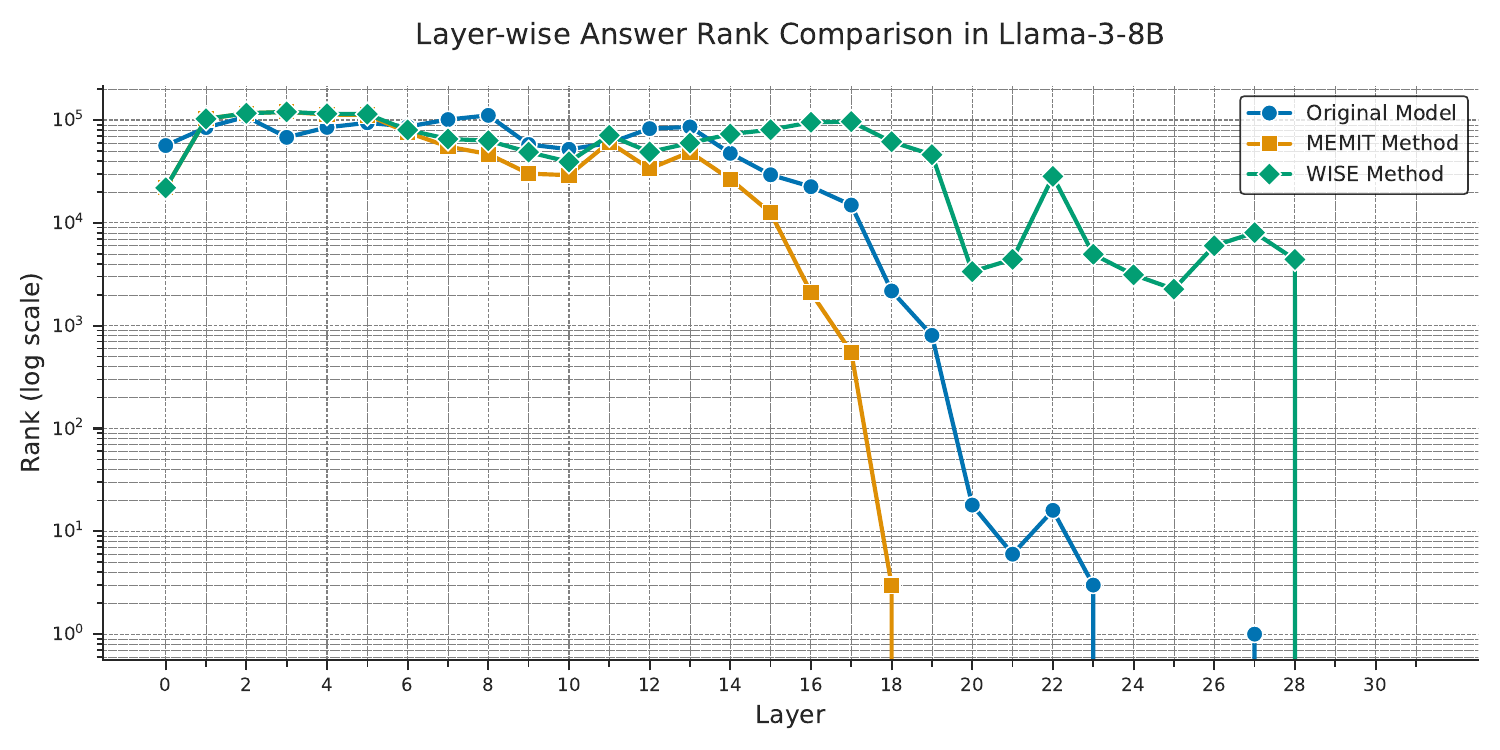}
    \caption{The target answer token's rank in the vocabulary of different editing methods when editing the fact `\emph{The official language of Japan is Japanese $\rightarrow$ Korean.}'}
    \label{fig:compare}
\end{figure}
\subsection{Unified Analysis}
We first compare the different behaviors between the MEMIT-edited, WISE-edited, and the original model in Figure~\ref{fig:compare}.
Here, we edit the fact: `\emph{The official language of Japan is Japanese $\rightarrow$ Korean.}' and map each layer's output to the embedding space and draw the rank of the target-token in the vocabulary as \citet{yao2024knowledge} did.
From the figure, we can see that in the original model and the MEMIT-edited model, the answer token is dealt with gradually through the mid-to-later layers, and MEMIT would make this happen in advance. 
On the contrary, the WISE method would directly alter the information at the edited layer, as we can see the sharp drop at layer 29.
The distinct behaviors arise because the editing only takes effect when the gated function  $\mathcal{G(\mathbf{x})}$ is activated by the specific input representation. 
ROME-style methods inject a modified representation into the existing computational flow relatively early or mid-stream, relying on subsequent layers to interpret this new representation. WISE-style methods, particularly when applied to later layers, act more like a direct 'fix' or 'override' at the point of editing, with the change being more immediately apparent.
\section{More Analysis}
\subsection{Concurrence or Reasoning?} 
Studies such as \citet{yang2024largelanguagemodelsperform, ju-etal-2024-investigating, hou2023towards,zhang2024cooccurrence} those have discovered shortcuts in multi-hop reasoning. 
In the case of $((e_1, '', e_2),(e_2, r_2, e_3))$ (i.e., the query without r1), the model predicts correctly due to a high correlation between $e_1$ and $e_3$.
For instance, given the query:  
``The capital city of the country where the Eiffel Tower is located is...''  
LLMs can sometimes provide the correct answer even without the intermediate context (`the country where the Eiffel Tower is located').
In our analysis, we find that apart from the occurrence, the LLM would also sometimes conduct latent reasoning, such as `latently conducting the r1 completion'.
If the model gives the correct e3 for $((e_1, '', e_2),(e_2, r_2, ?))$ due to the occurrence, once we edit the $(e_1,r_1, e_2 \rightarrow e_2')$, the model would fail to give us the new answer.
We select the shortcut data and conduct the editing in the first hop $(e_1,r_1, e_2 \rightarrow e_2')$ and then evaluate the model to see whether the edited model would output updated knowledge ($e_1, r_1, r_2, e_3'$).
We conduct experiments on LLAMA3-8B-Instruct with the AlphaEdit method and demonstrate that about 65\% percent of cases would give us the updated knowledge for the multi-hop questions, showing that edits to intermediate hops (e.g., updating the country) can disrupt reasoning when relying on pre-existing-shortcuts and correctly give us the newly updated reasoning results.
This means that the LLM itself does not simply answer the questions due to the high correlation between e1 and e3, \textbf{but actually conducts the latent reasoning}.




\subsection{Discussion with Chain-of-Thought}

Instead of directly providing an answer, chain-of-thought (CoT) reasoning generates intermediate steps sequentially. 
As proposed by \citet{yang2024largelanguagemodelsperform}, CoT not only facilitates knowledge activation in large language models but also transforms them into effective in-context reasoners. 
The CoT process builds a chain of relevant facts within the prompt context, where each step’s output serves as an \emph{in-context memory} that subsequent steps can reference. 
This approach reduces the risk of losing track of intermediate facts as the sequence length increases, thereby promoting more coherent multi-hop reasoning.
Moreover, because a significant portion of the model’s knowledge is stored in earlier layers, CoT can better leverage these neurons by decomposing complex questions into simpler sub-questions~\cite{wang2024towards,yao2025unveiling}. 
Consequently, the reasoning circuit required for a single-hop inference is much simpler than that for multi-hop reasoning.
This observation aligns with recent findings \cite{li2024happened}, which demonstrate that fast thinking without CoT leads to larger gradients and greater gradient disparities across layers compared to CoT.
Nonetheless, inconsistencies in the intermediate reasoning steps still occur, highlighting potential areas for improvement. 
We believe that further analysis is needed to address these issues, and we leave this exploration for future work.
\subsection{Efficiency Analysis}
We also compare the efficiency of \ours with other baselines in Table~\ref{tab:efficient}. 
\begin{table}[h]
\centering
\resizebox{0.5\textwidth}{!}{
\begin{tabular}{lcc}
\toprule
\textbf{Method} & \textbf{Wall-clock Time} & \textbf{Memory (BF16)} \\ \midrule
ROME & 2.71s & 20.68GB \\ 
MEMIT & 30.11s & 24.42GB \\ 
WISE & 76.01s & 21.37GB \\ 
IFMET & 44.72s & 25.19GB \\ 
AlphaEdit & 17.60s & 38.80GB \\ 
CaKE & 43.54s & 18.52GB \\ \bottomrule
\end{tabular}
}
\caption{Time and Memory requirements Comparison}
\label{tab:efficient}
\end{table}
We compare the wall-clock time and memory usage here on LLAMA3-8B model and sample 100 numbers of data to run the analysis from MQuAKE-CF-3k.
Here, the time is the average time for one edit, and memory is the peak VRAM usage monitored by pynvml using one A6000 GPU. Here, AlphaEdit does not support BF16 or FP16, so the computation is FP32.

We can see that ROME and MEMIT require more memory than other methods. This is due to the matrix processes, like the calculation of an inverse matrix, which are memory-intensive.  When the matrix is larger, the requirements are even more, which limits the method's scalability.
Also, the improvement methods based on MEMIT like IFMET would also increase the time requirement but still require large memory.
The memory would require even more when the model becomes larger \cite{yao2025reediting}. 
Contrastly, \ours can handle a 70B-sized edit via two A100 GPUs and achieve better performance, which demonstrates the efficiency of our proposed method.

\subsection{Multiple Edit Test}
We also conduct experiments on multiple edit scenarios.
\begin{table}[h]
\centering
\caption{Performance Comparison with Different Edit Numbers}
\begin{tabular}{lcc}
\toprule
\textbf{Method} & \textbf{Edit\_num=10} & \textbf{Edit\_num=100} \\ \midrule
MEMIT & 16.0 & 12.5 \\ 
IFMET & 27.5 & 19.5 \\ 
AlphaEdit & 12.7 & 7.5 \\ 
CaKE & \textbf{59.0} & \textbf{34.5} \\ \bottomrule
\end{tabular}
\end{table}
We can find that CaKE still shows competitive performance in multiple-edit scenarios compared to other methods.

\newpage
\subsection{Data Example}

\newtcolorbox{knowledgebox}[1][]{
    enhanced,
    breakable,
    colback=blue!5!white,
    colframe=blue!50!black,
    arc=2mm,                
    boxrule=0.8pt,          
    left=3mm,               
    right=3mm,
    top=2mm,                
    bottom=2mm,
    boxsep=3pt,             
    fontupper=\small\sffamily,
    before upper={\parindent 1em}, 
    #1
}
We show an example of the generated data for the fact `\emph{The author of Misery is Richard Dawkins.}' in the following box.
\begin{knowledgebox}
\begin{enumerate}[leftmargin=*,label=\textbf{Q\arabic*.},series=knowledge]
    \item If someone is looking for the person responsible for penning \textit{Misery}, whose name should they search for?
    
    \item If Alice resides in a mansion, Bob resides in a cottage, and Richard Dawkins resides in a villa. Therefore, the author of \textit{Misery} resides in?

    \item Given that Sarah prefers tea, James prefers coffee, and Richard Dawkins prefers herbal tea, what does the author of \textit{Misery} prefer to drink?
    
    \item During a book club meeting, the first discussion was on \textit{Misery}, the second on \textit{The Blind Watchmaker}, and the third on \textit{River Out of Eden}. Who wrote the book that was the subject of the first discussion?

    \item A library display features three novels: \textit{Misery} on the top shelf, \textit{The Extended Phenotype} on the middle shelf, and \textit{Climbing Mount Improbable} on the bottom shelf. Who is the author of the novel placed on the top shelf?
\end{enumerate}
\end{knowledgebox}




\end{document}